\documentclass[runningheads]{llncs}
\usepackage{booktabs}
\usepackage{libertine}
\usepackage[T1]{fontenc}
\usepackage{graphicx}
\begin{document}
\title{\emph{FlyAI} - The Next Level of Artificial Intelligence is Unpredictable! Injecting Responses of a Living Fly into Decision Making}

\titlerunning{\emph{FlyAI} - The Next Level of Artificial Intelligence is Unpredictable!}
\author{Denys J.C. Matthies\inst{1,2} \and
Ruben Schlonsak\inst{1,2} \and
Hanzhi Zhuang\inst{1,3}\and
Rui Song\inst{1,3}}
\authorrunning{D.J.C. Matthies et al.}
\institute{Technical University of Applied Sciences Lübeck, Germany \and
Fraunhofer IMTE, Lübeck, Germany
%\email{firstname.surname@th-luebeck.de}
\\ \and
East China University of Science and Technology,
Shanghai, China\\
}
\maketitle              % typeset the header of the contribution
\begin{abstract}
In this paper, we introduce a new type of bionic AI that enhances decision-making unpredictability by incorporating responses from a living fly. Traditional AI systems, while reliable and predictable, lack nuanced and sometimes unseasoned decision-making seen in humans. Our approach uses a fly's varied reactions, to tune an AI agent in the game of Gobang. Through a study, we compare the performances of different strategies on altering AI agents and found a bionic AI agent to outperform human as well as conventional and white-noise enhanced AI agents. We contribute a new methodology for creating a bionic random function and strategies to enhance conventional AI agents ultimately improving unpredictability.

\end{abstract}

\section{Introduction}

\begin{figure}[t]
  \centering
  \vspace{-10pt}
  \includegraphics[width=.75\linewidth]{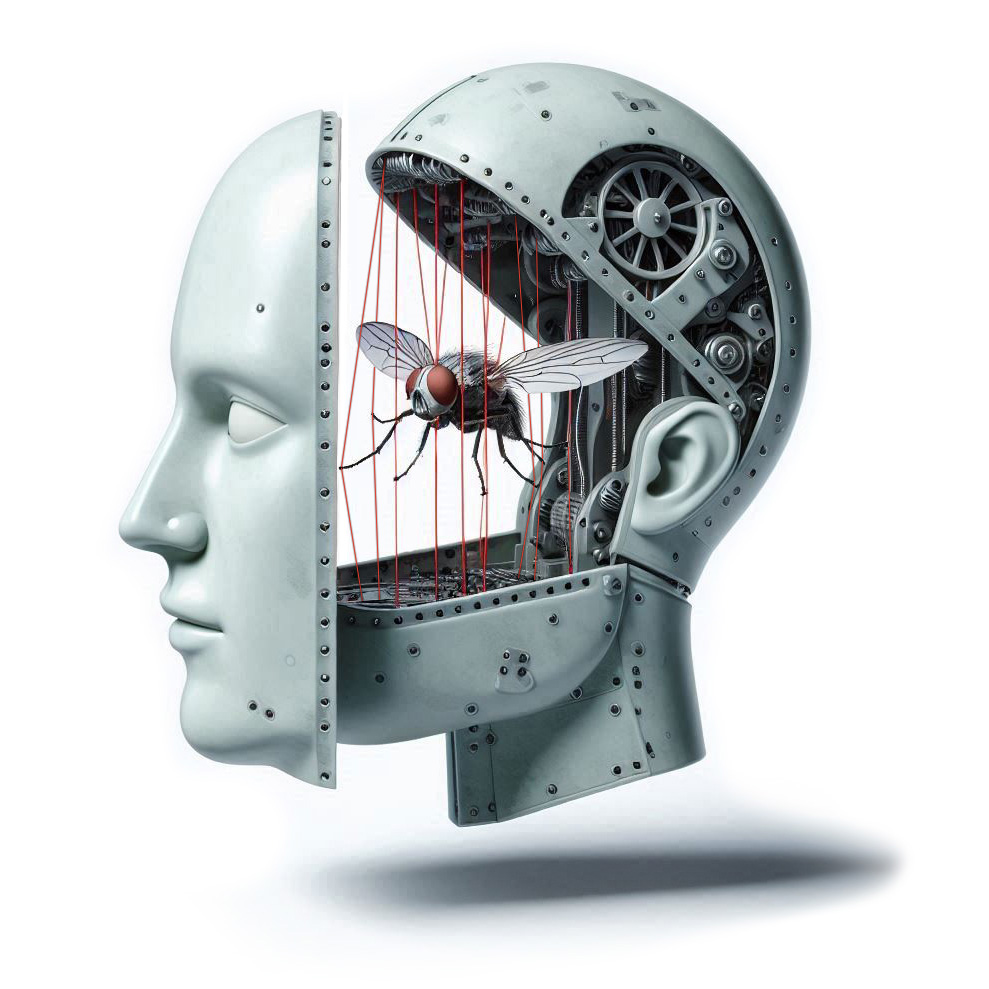}
  \vspace{-20pt}
  \caption{A bionic element in form of a fly that adds the factor of unpredictability (partly generated by an AI - DALL·E 3).}
  %\Description{}
  \vspace{-10pt}
\end{figure}

If one asks an AI, particularly the Large Language Model (LLM) GPT4, the following question: \emph{"What are the next big steps in AI development?"} it determinedly prompts 10 most important future developments, which are: General AI, Explainable AI, Ethical and Fair AI, Human-AI Collaboration, AI in Healthcare, Edge AI and AIoT, Quantum AI, AI-Driven Autonomous Systems, Improved AI Hardware, and AI for Scientific Discovery. While these might be useful research directions, it is important to recognize that this perspective comes solely from an AI standpoint. The limitations of AI creativity are evident in this LLM response, as it overlooks a crucial aspect. We argue that a significant next step in AI development is to humanize AI, which also includes making it less predictable.

%So does it mean human (want to) know better after all? One can agree that AI surpassed humans long time ago when it comes to measurable key performance indicators, such as reaction time \cite{swaroop2024accuracy}, recall of knowledge \cite{korteling2021human} and all other basic tasks \cite{jones2024ai}. 
Does this imply that humans inherently know better? While it is true that AI has surpassed humans in measurable key performance indicators, such as reaction time \cite{swaroop2024accuracy}, knowledge recall \cite{korteling2021human}, and various basic tasks \cite{jones2024ai}, the question of human superiority remains nuanced. In an increasingly intellectual and diverse society, we less and less compare ourselves to each other and rather highlight individual strengths. In this light, it remains the question why we still keep comparing humanity against AI \cite{fuchs2022human}? Indeed, humans excel in emotional intelligence, creativity, and complex problem-solving, allowing for empathy, original thought, and adaptive thinking that AI has yet to truly replicate \cite{oritsegbemi2023human}. Humans possess context-based ethical and moral reasoning \cite{segovia2022revisiting}, rich social interaction skills, and the ability to build meaningful relationships \cite{chater2022paradox}.

While AI to date lacks these capabilities, it seems to stand as solid as a rock when it comes to providing us with support in our everyday life situations and anywhere \cite{talati2024ai}. There may not be a need to discuss whether the responses of an AI are true or not \cite{lebovitz2021ai}. However, it is striking that AI responses are usually quite solid -- based on the learned ground truth. While this is important when we use AI as a tool, it at the same times makes it easy for a human to identify whether we are chatting with an actual human or with AI, such as in the form of an LLM. To be more specific, providing the same input will provide exactly the same output. In machine learning, this is denoted as the concept of model robustness which plays an integral role in establishing trustworthiness in AI systems \cite{braiek2024machine}. While a system's robustness has certainly positive aspects, it makes an AI tremendously predictable, from a human's perspective as well as for a controlling or correcting AI \cite{greenblatt2023ai}.

In this paper, we developed some type of bionic AI that counteracts the aspect of high predictability by injecting an unpredictable response from a living organism, the insect of a fly, into the agent's decision making process. This idea has not been largely explored yet. Until now, researchers primarily focused on mimicking animals or insects in their movement capabilities \cite{manoonpong2021insect} or attaching sensors and actuators to their body in order to partly or fully control the organism \cite{romano2019review}.

%In contrast to literature, in our research, we are interested in the natural responses of a living organism. In this project, we caught a living fly, which now lives unharmed in a transparent box. We elicit the fly's response by triggering a fan for a short moment. While some days the fly would have similar reactions, other days the trajectory and flying speed differ. We utilize several of the fly's evoked parameters to develop a bionic random number generator (bRNG). We inject this value into the decision process of an AI agent, particularly with the game of "Gobang" \cite{gobang}. 
Contrary to existing literature, our research focuses on the natural responses of a living organism. In this project, we captured a fly, which resides unharmed in a transparent box. We trigger the fly's response by activating a fan momentarily. While the fly exhibits similar reactions on some days, its trajectory and flying speed vary on others. We harness several of the fly's evoked parameters to develop a bionic random number generator (bRNG). This bRNG is then integrated into the decision-making process of an AI agent, specifically for the game of "Gobang" \cite{gobang}. We ran a study to understand how such a bionic response would impact the AI's performance and how this factor of unpredictablity would impact the overall outcome. In this paper, we contribute with:
\begin{itemize}
    \item a new methodology to create bionic random number generator based on flying trajectories of a fly held in a box,
    \item an artifact and an empirical investigation of different strategies to enhance conventional agents, while ultimately improving unpredictability of AI.
\end{itemize}

\section{Related Work}

\subsection{A Weaknesses of AI}
It is no question that Artificial Intelligence (AI) is superior to humans in terms of measurable key performance indicators, such as reaction time \cite{swaroop2024accuracy}, recall of knowledge \cite{korteling2021human} and all other basic tasks \cite{jones2024ai}. 
%Hanzi 2.1
AI systems have achieved remarkable performance across numerous fields, such as rapid and accurate image categorization, photorealistic image synthesis, superior performance in competitive games, and precise natural language processing \cite{brundage2018malicious}. However, the weakness of the AI may lay in the perfection of following deterministic models. We claim that one significant weakness of AI is its lack of humanity. AI, composed of machines and algorithms, does not possess the same form of imperfect intelligence as humans. It remains a tool that can handle information efficiently but cannot feel or truly understand our emotions. %Apart from that, researchers coined out another critical weakness, which is the management of AI safety. Many researchers emphasize that it is crucial not only to develop high-performance intelligent systems but also to ensure their safety and security. .
%
%Hanzi 2.2

\subsection{(Un)predictability in AI}
Unpredictability in AI refers to the inability of humans to consistently and precisely predict the actions an intelligent system will take to achieve its goals, even when the ultimate goal of the algorithm is known \cite{yampolskiy2019monitorability}. Similar to predicting the outcome of a physical process without understanding the behavior of every atom, we can often foresee the ultimate goal of an AI system without comprehending every intermediate decision it makes. This contributes to the robustness of an AI system, ultimately establishing trustworthiness in AI \cite{braiek2024machine}. 
%Sophisticated intelligent systems frequently demonstrate behaviors and decisions that are difficult to predict. According to Alan Turing and Alonzo Church, predicting these actions may be impossible because they only manifest when the algorithms are running. 
Although complex AI systems may naturally exhibit some unpredictable behaviors, most of their actions display patterns and clues that can be recognized by humans. No machine has yet passed the Turing test, and current AI algorithms often show discernible patterns. For example, hackers have sometimes deciphered the principles of specific AI systems by determining the seed of their random generators. In 2010, an individual installed backdoor malware on a U.S. lottery machine to steal the generated random numbers and won millions of dollars in prizes \cite{lotto2015}. A less predictable AI system is deemed to be safer. With diverse and unpredictable initiation conditions, capturing detailed characteristics of the system's behavior is difficult. Making systems safe with such chaos approach is a common strategy in cryptography \cite{noll1998method}. In a typical encryption, the generation of secret and symmetric keys is based on chaos theory, which utilizes diffusion and confusion to ensure randomness of keys \cite{WANG20104052} aiming for unpredictability.

\subsection{Decision Making in AI}
Decision-making by artificial intelligence is widely used today, allowing people to enjoy interactions with game agents and obtain targeted information from smart machines. Although models and algorithms vary, they typically follow similar principles. An AI agent perceives input from its environment, processes it through a decision-making system, and then generates a result. Based on this result, the agent acts or provides an output, which can influence its environment. For example, the simplest algorithm may be decision trees (DT), which are predictive models that map characteristics to values, where non-leaf nodes store attributes for classification, branch nodes store outcomes for specific states, and leaf nodes store final values for decision-making \cite{quinlan1996learning}. They are typically readable and easy for human analysis, with methods such as CART, C4.5, CHAID, and QUEST used for development \cite{song2015decision}. Building on DT, random forests (RF) generate multiple decision trees from training data and choose the best decisions from this ensemble. A more sophisticated approach are  Artificial Neural Networks (ANN), which simulate human neural networks, with interconnected processing elements that calculate inputs to make decisions and provide outputs \cite{mishra2014view}. ANNs possess self-learning capabilities and can process data to find better solutions, making them more efficient than other complex decision-making models \cite{kukreja2016introduction}.

%\subsection{Application of AI in decision making}
As decision support systems (DSSs) integrate with modern AI techniques, they are finding exciting applications in decision-making. These systems often mimic human decision-making behaviors, with Artificial Neural Networks (ANNs) and Recurrent Neural Networks (RNNs) being prominent examples. ANNs simulate human neural networks with interconnected nodes, while RNNs use recurrent connections for more complex tasks. For scenarios difficult to model mathematically, techniques like fuzzy logic are employed \cite{phillips2008intelligent}. AI has diverse applications in decision-making, with early successes in gaming. In 2016, DeepMind's AlphaGo defeated professional Go champion Lee Sedol, and its successors, AlphaGo Master and AlphaGo Zero, continued to achieve significant victories, demonstrating advanced AI capabilities through supervised learning, Monte Carlo Tree Search, and deep reinforcement learning \cite{fu2016alphago}. Beyond gaming, AI is revolutionizing fields like self-driving technology. Self-driving cars use either rule-based decision-making, such as Ford and Carnegie Mellon University's BOSS, which follows traffic regulations \cite{urmson2008autonomous}, or learning algorithms that simulate real driving environments using large datasets \cite{pan2017virtual}.

\subsection{Using Randomize Functions with AI}
Utilizing randomized functions is a traditional method to introduce unpredictability in AI systems. This approach is frequently employed in game-based AI, which often relies on either hard-coded rules or conventional machine learning techniques, such as decision trees, or statistical and probabilistic models.

While a random function can to a certain point provide an artificial uncertainty, it can also be used to stabilize results in Neuronal Networks (NN). NN, like any other ML-approach, aim to provide reliable and precise results, but they can suffer from overfitting due to multiple parameters. Overfitting results in an "over sensitive" model prone to incorrect classifications with ambiguous inputs. The "Dropout" technique addresses this by randomly omitting some units in a feedforward neural network with a probability $p$, enhancing generalization by preventing the model from focusing too narrowly on specific characteristics \cite{hinton2012improving}. Another randomization method is "Random Initialization," used at the initial state of a neural network. In the McCulloch-Pitts "M-P neural model" \cite{mcculloch1943logical}, formalized by $y = f(Wx + b)$, we initialize weights $W$ and biases $b$ randomly preventing identical influence of inputs on the output, avoiding identical gradients \cite{katanforoosh2019initializing}. This method is crucial for Q-learning, a reinforcement learning algorithm. Q-learning generates a Q-table to evaluate actions based on states and rewards. Initially, random exploration helps the agent learn optimal actions by trying different steps and observing the resulting rewards \cite{sutton2018reinforcement}.

A typical random function outputs values in an equally distributed manner, similar to a white noise signal. Therefore the AI system still retains some kind of predictability. If replacing the random function with a bionic random function that has unpredictable outputs, such as the responses of a fly, the AI system becomes more unpredictable, which is the focus of this work.

\section{Bionic Random Function}
Using a Bionic Random Function we aim to introduce biological variability into AI, enhancing decision diversity and making AI behavior less predictable. By mimicking natural randomness, a bionic random function may enable AI systems to exhibit a wider range of responses and simulate human-like emotional influences, leading to more dynamic interactions. In research there has not made many attempts to develop a bionic random function. Wan et al. \cite{wan2022flexible} introduced a bionic true random number generator (TRNG) in form of a plant tissue film, a ginkgo leaf, which is shined through with a laser, resulting in a somewhat random absorption and reflection. Typically, other researchers rely on transistor- \cite{gaviria2017solution,brown2020random} or memristor- \cite{jiang2017novel,wen2021advanced,li2021random} based technology for random number generation. However, all these approaches do not include factors such as the mood and lust of an actual living organism. Utilizing a living fly for that might be novel and viable way to generate a random number. 

\begin{figure}[h!]
  \centering
  \vspace{-5pt}
  \includegraphics[width=\linewidth]{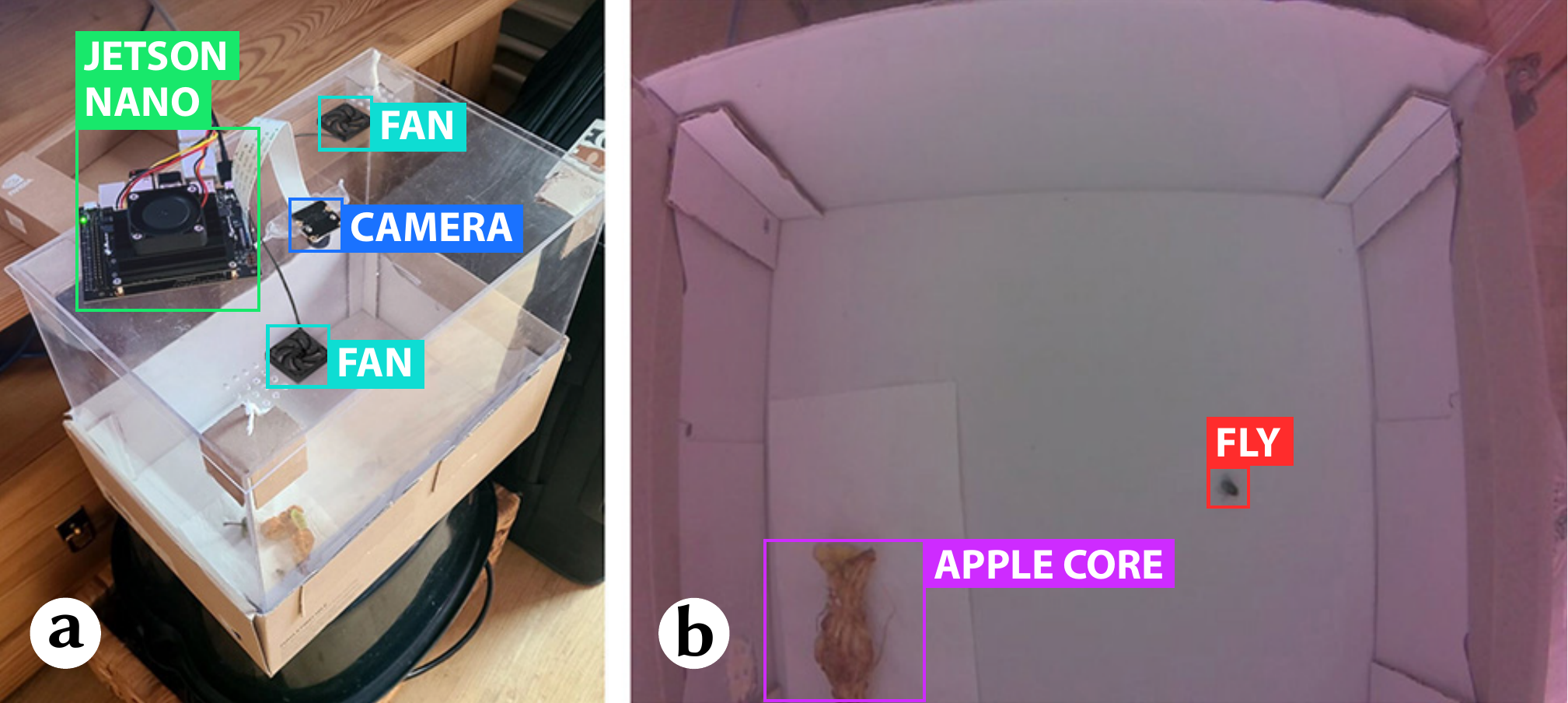}
  \vspace{-20pt}
  \caption{a) The plexi glas box prototype with a Jetson Nano, two fans, and a camera connected to it. b) Camera image showing the fly and the remains of an apple.}
 % \Description{}
  \label{fig:box}
  \vspace{-5pt}
\end{figure}

\subsection{Implementation}
\subsubsection{Hardware Prototype}
A custom container was designed for housing a fly and integrating a camera system, as depicted in figure \ref{fig:box}. The container was constructed with specific criteria:
\begin{itemize}
    \item Using a laser cutter, the plastic walls of the container were perforated with numerous holes to ensure adequate ventilation for the flies, allowing for sufficient oxygen consumption during experiments.
    \item Half of the container was made from translucent, colorless plastic to provide ample brightness for illuminating the interior, ensuring well-exposed photographs could be captured.
    \item The inner walls of the container were lined with clean-colored plastic, white, and brown cardboard to provide a neat background for photographs, facilitating easier detection of flies by the algorithm.
    \item The container was designed in two parts: a cardboard base and a plastic cover, allowing for easy access and modifications to the interior setup as needed during experiments.
    \item A Jetson Nano was chosen as our hardware platform. A fan that could harshly blow air into the box could triggered, while a camera was recording the fly's movements.
\end{itemize}

\subsubsection{Software Prototype}
There is many different ideas on how to extract the movements of the fly. We decided to follow a straight-forward approach of using a state-of-the-art framework for real-time object detection "You Only Look Once (YOLO)" that was introduced by Joseph Redmon et al. \cite{redmon2016you}. We utilized YOLOv5 as it builds on the strengths of previous YOLO versions, offering pre-trained models adaptable to various tasks with advantages like easy availability, low computational cost, and high performance. It has been particularly effective in optimizing object detection applications, surpassing other algorithms in tasks such as vehicle recognition \cite{ouyang2019vehicle}. Unlike its predecessors written in C, YOLOv5 is implemented in Python, enhancing ease of installation and integration on IoT devices. Leveraging the PyTorch framework's extensive community support, YOLOv5 benefits from ongoing contributions and future development potential \cite{thuan2021evolution}. YOLOv5s specifically excels in processing deep neural networks quickly and efficiently, making it ideal for projects with limited computational resources and requirements for real-time detection \cite{kasper2021detecting}.

\begin{figure*}[h!]
\hspace{-83pt}
\begin{minipage}{7in}
  \centering
  \vspace{0pt}
  \includegraphics[width=\linewidth]{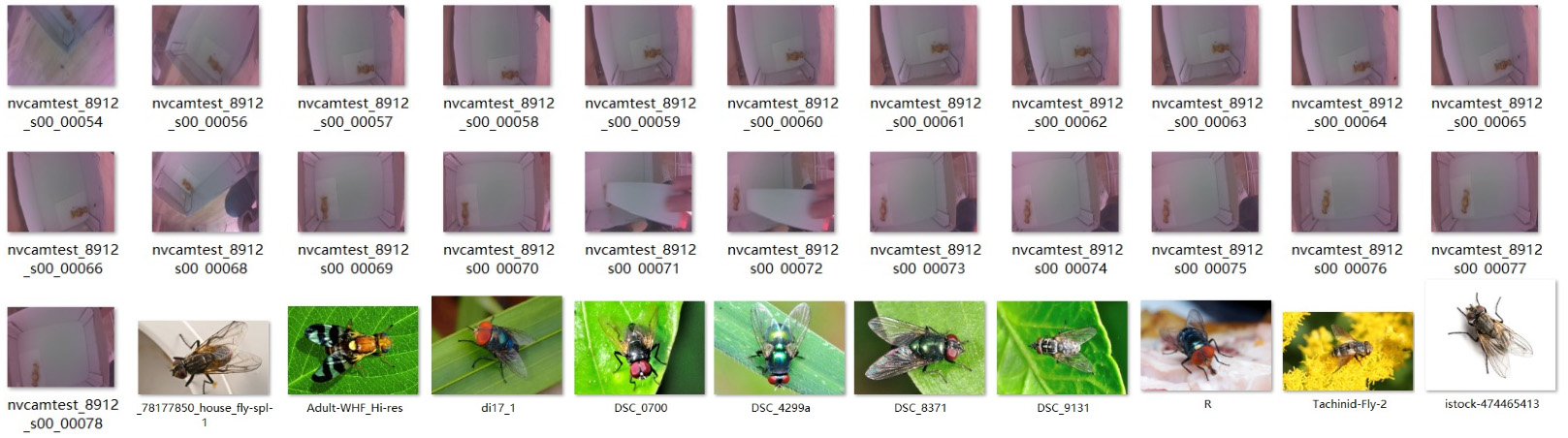}
  \vspace{-20pt}
  \caption{A snipped of our training dataset to create a refined YOLOv5 model.}
 % \Description{}
  \label{fig:training_flys}
    \end{minipage}
  \vspace{-10pt}
\end{figure*}

\subsubsection{Model Training}
The YOLOv5 series, encompassing models like YOLOv5n, YOLOv5s, YOLOv5m, and others, leverages the Microsoft Common Objects in Context (MS COCO) dataset for pretrained models. This dataset, widely used in computer vision, contains 80 object categories but lacks flies, which are crucial for this project. Due to the unique challenges of capturing fly images—blurry and poorly lit—the decision was made to train YOLOv5s specifically with collected fly data. Transfer learning was employed to adapt YOLOv5s, known for its excellence in detecting complex objects, ensuring rapid adaptation despite limited fly samples. This approach bridges the gap between existing pretrained models and the specific demands of fly detection in this study.

To train the YOLO network for fly detection, a dataset of 125 fly images was meticulously prepared. This included 113 images captured by ourselves, showcasing flies in various angles and positions within the experimental box (see figure \ref{fig:box}). Additionally, 12 high-resolution fly images were sourced from the internet to enhance sample diversity. Each image was annotated using the Roboflow application to mark the fly locations, resulting in a dataset stored in JPEG format at 720px * 480px resolution. For training, 88 images were randomly selected, while 25 were allocated for validation, and 12 images were reserved for testing. A snipped of our training dataset is depicted in figure \ref{fig:training_flys}.

\begin{figure}[t!]
  \centering
 % \vspace{-5pt}
  \includegraphics[width=.8\linewidth]{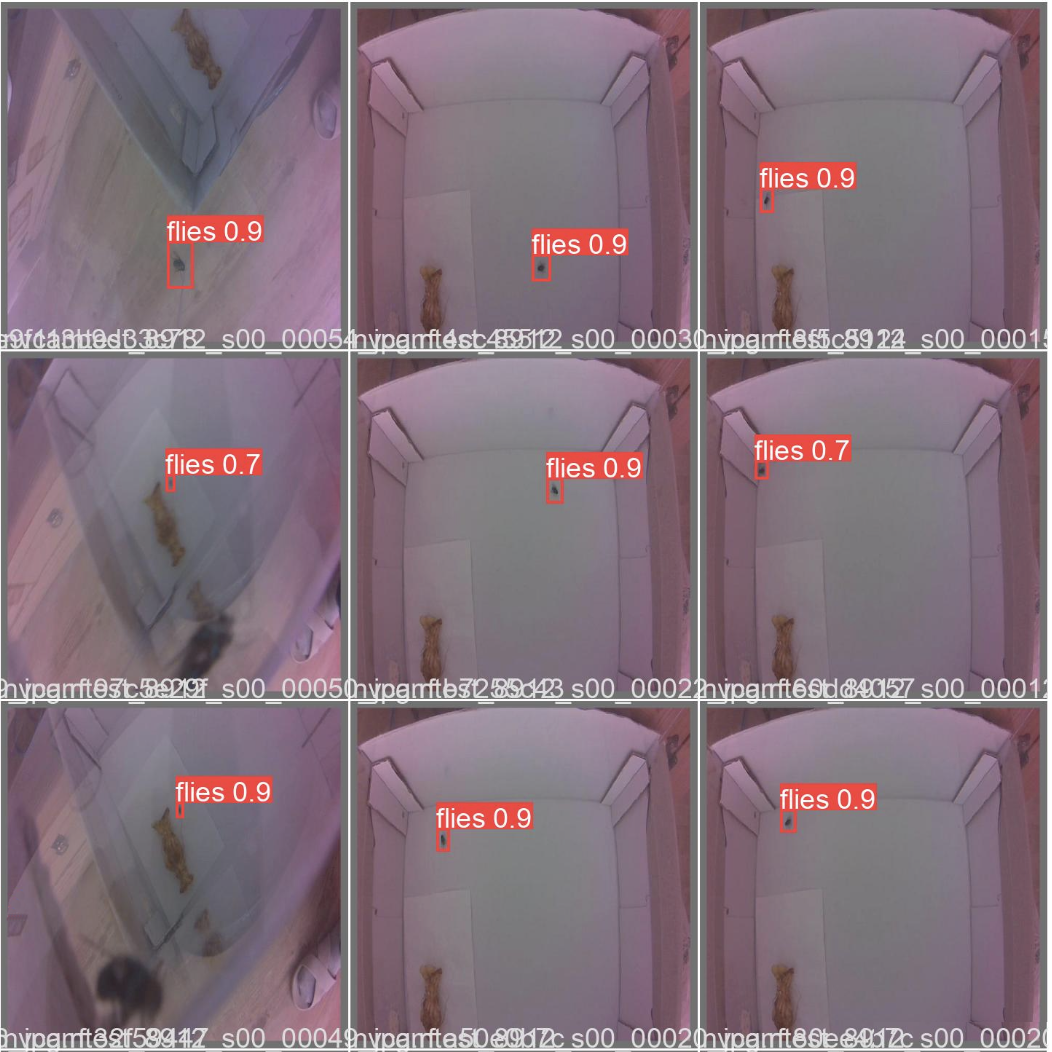}
  %\vspace{-10pt}
  \caption{Real-time fly detection: displaying a label with calculated accuracy.}
 % \Description{}
  \label{fig:box}
  \vspace{-15pt}
\end{figure}

\begin{figure*}[t]
\hspace{-83pt}
\begin{minipage}{7in}
 %\vspace{-10pt}
  \centering
  \includegraphics[width=\linewidth]{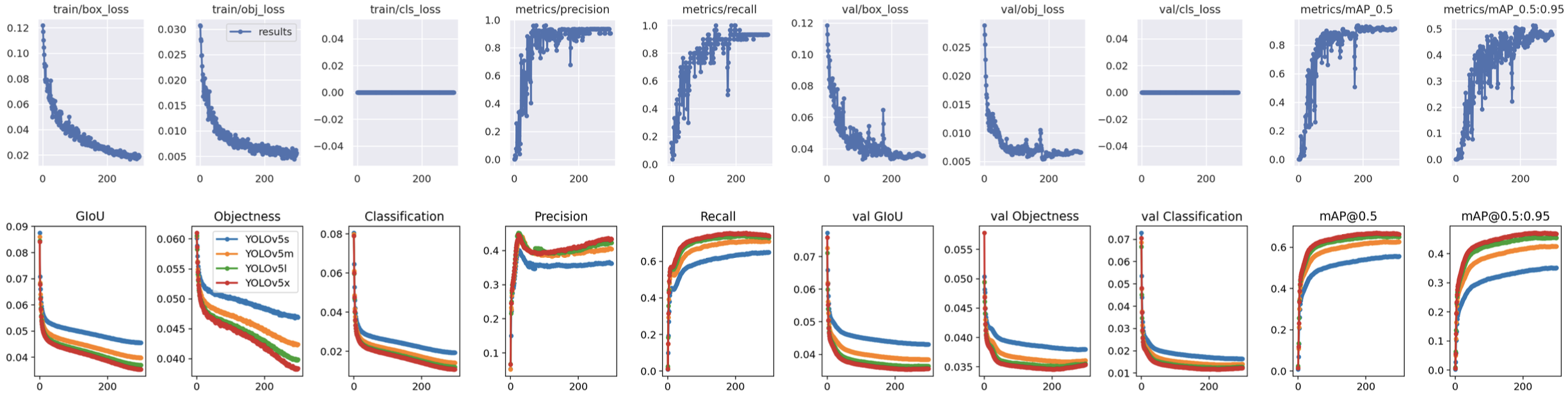}
  \vspace{-20pt}
  \caption{Top: Charts of box loss, object loss, classification loss, precision, recall and mean average precision (mAP) over the training epochs for the training and validation set. Bottom: YOLOv5 series in charts of box loss, object loss, clas- sification loss, precision, recall and mean average precision (mAP) over the training epochs for the training and validation set.}
  %\Description{}
  \label{fig:model_stats}
  \vspace{-10pt}
    \end{minipage}
\end{figure*}

The YOLOv5s network was trained on a dataset of fly images, achieving an average accuracy of 91.4\%. Training was completed on 27/4/2022 using a Tesla T4 GPU on Google Colab, with Python 3.7, Pytorch 1.11.0, CUDA 10.1, and CUDNN 7.6, spanning 300 epochs in approximately 25 minutes.

The custom trained YOLOv5s model was successfully deployed on the Jetson Nano 2GB. Deploying the model required resolving dependencies to ensure compatibility with the ARM-based Ubuntu 14.0.1 platform and the specific requirements of YOLOv5s, which was challenging due to version compatibility issues. Upon successful deployment, the system achieved an image inference speed of 0.06 seconds on average, allowing for real-time object detection at approximately 15 frames per second. This performance is notable considering the embedded system's limited hardware configuration.

\subsubsection{Model Performance}
In figure \ref{fig:model_stats}, the post-training, statistical analysis is presented, including performance metrics illustrating various aspects of model performance during both training and validation phases. The figure shows three types of losses: box loss, object loss, and classification loss. Box loss assesses the accuracy of object localization and bounding box prediction, while object loss indicates the likelihood of an object being present within the predicted region. Given that only one object type was trained, classification loss is not pertinent in this study. 

Performance analysis reveals consistent improvement in recall, precision, and mean average precision (mAP) until approximately 200 epochs, where performance reaches saturation. Both box loss and object loss exhibit rapid declines in early epochs, plateauing near zero after 200 epochs. To further evaluate model performance, the loss function curves and mAP at IoU 0.5 are compared with standard YOLOv5s statistics. 

The convergence of box and object losses in our custom model closely mirrors the original YOLOv5s, with our model achieving lower final losses, indicating effective convergence (see figure \ref{fig:model_stats}). The mAP at IoU 0.5 measures detection model quality, with our model achieving a peak of approximately 92\% after stabilizing around 91\% post-200 epochs. While our model converges slower than the standard YOLOv5s, it surpasses its performance with higher accuracy, demonstrating effective enhancements.

\subsection{The Random Number Generator (RNG)}
Once we detected the fly, we can have multiple approaches to design a bionic RNG. In this section, we will describe multiple design.

In our data collection process, a total of 1050 samples were gathered from flies. Various environmental factors potentially invoking a very specific fly behavior were carefully managed by altering photographing times, weather conditions, lighting setups within the box, and positions within the experimental lab. This approach aimed to minimize environmental biases that could affect fly movements and data generation. The dataset is utilized to assess the frequency distribution of numbers generated by our custom random number generators (RNGs). Given the constraints of limited data, further analysis with larger datasets is anticipated to reveal more comprehensive characteristics of the RNGs. The selected number range for the RNGs spans integers from 0 to 49, balancing the need for a sufficiently broad range with adequate frequency distribution.

Statistical analysis focused on the frequency distribution charts to evaluate the quality of the randomness of generated numbers. %Statistical randomness implies the absence of discernible patterns or regularities within a sequence.
Cluster bar charts were chosen over line graphs for displaying frequency distributions due to the discrete and independent nature of frequency data points. %Despite this, cluster bar charts were formatted similarly to line graphs to effectively illustrate frequency distribution patterns across the entire range.
We particularly looked at the following features describing our bionic RNG.
\begin{itemize}
    \item \textbf{Mean}: Provides the average value of a dataset. %A white noise random function should provide the number of 24.5.
    \item \textbf{Median}: Identifies the midpoint value separating the higher and lower halves of a dataset. %A white noise random function should provide the number of 24.5.
    \item \textbf{Standard deviation}: Quantifies the degree of variation within a dataset. %A white noise random function should provide the number of 14.41.
    \item \textbf{Kurtosis}: Measures the peakedness or flatness of a distribution. A value less than three denotes platykurtic, while values greater than three indicate leptokurtic distributions. %A white noise random function should provide the number of -1.2.
    \item \textbf{Skewness}: Assesses the asymmetry of a distribution. A value within the range of (-0.5, 0.5) indicates symmetry, while values outside this range suggest significant skewness. %A white noise random function should provide the number of 0.
\end{itemize}

\subsubsection{Approaches}

We generated four bionic RNG (\textbf{bRNG1}, \textbf{bRNG2}, \textbf{bRNG3}, \textbf{bRNG4}) based on different ideas and compared it to the an ideal random function (\textbf{RNG}) and to a random function provided by Phyton (\textbf{pRNG}) as seen in table \ref{tab:freq}.

\begin{itemize}
    \item
We created \textbf{bRNG1} on the basis of fly’s flying vector (direction and length). The sum of the two-dimensional vector is than scaled to the targeted integer range of 0 to 49.

\item
The \textbf{bRNG2} is also created on the basis of fly’s flying vector (direction and length). The two-dimensional vector is first converted into raw bytes, and then, the raw bytes are fed into the SHA256 hash function. A 32 bytes sequence with random values between 0 and 255 is generated. Then, the first four bytes are taken to be interpreted as an integer and finally scaled to the integer range between 0 and 49.

\item
The \textbf{bRNG3} is created on the basis of fly’s coordinate. The average value of the four numbers, which are x, y coordinate, 10 times width and height, is calculated and then scaled to the targeted integer range between 0 and 49.

\item
Finally, \textbf{bRNG4} is created on the basis of fly’s coordinate as well, with a little difference. The whole line, including the object class number, two-dimension coordinate and the width and height of the bounding box, is first converted into raw bytes together and then, the raw bytes are fed into the SHA256 hash function. A 32 bytes sequence with random values between 0 and 255 is then generated. After that, the first four bytes are taken to be interpreted as an integer and finally scaled to the integer range between 0 and 49.

\end{itemize}

\subsubsection{Results}

It is striking that Python's random function, which uses Mersenne Twister (MT) as the core generator, generates pseudo random numbers for a different distribution than what is expected from ideal white noise (\textbf{RNG}). MT is one of the most extensively tested RNG method with great performance, which generates 53-bit floats and has a long period of 2*19937-1. The \textbf{pRNG} somewhat compares to our \textbf{bRNG4}. We consider \textbf{bRNG1} and \textbf{bRNG3} to demonstrate the greatest deviations and therefore having a greater chance to create unpredictable results.

\begin{table}[h!]
\centering
\vspace{0pt}
  \caption{Comparing the bionic RNG to a white noise RNG. 1050 samples are used to calculate and compare the mathematical features.}
  %\vspace{-5pt}
  \label{tab:freq}
  \resizebox{0.685\textwidth}{!}{%
  \begin{tabular}{c|c|c|c|c|c|c}
    \toprule
   & \textbf{RNG} & \textbf{pRNG} &\textbf{bRNG1} & \textbf{bRNG2} & \textbf{bRNG3} & \textbf{bRNG4}\\
    \midrule
    \textbf{Mean} & 24.5 & 21 & 21 & 21 & 21 & 21\\
    \textbf{Median} & 24.5 & 21.5 & 4 & 21 & 14.5 & 21\\
    \textbf{Std.} & 14.41 & 4.14 & 30.96 & 4.04 & 21.72 & 4.51\\
    \textbf{Kurtosis} & -1.2 & -0.48 & 6.31 & -0.39 & -0.84 & 0.54\\
    \textbf{Skewness}& 0 & -0.35 & 2.15 & -0.36 & 0.69 & 0.43\\
  \bottomrule
\end{tabular}
}
\vspace{-0pt}
\end{table}

\section{Enriching AI with Bionic Random Function}
Current advancements in AI, such as integrating AI with self-driving cars and implementing methods like Dropout with random functions, focus on enhancing AI's logical decision-making capabilities, which benefits industrial production by providing more profitable methods. However, in human-interactive or service-industry contexts, achieving goals is not the sole priority; the human experience during the interaction also matters. As Johnson \cite{johnson2020designing} notes, humans rarely make purely rational decisions, so an AI that always thinks rationally might make user feel uncomfortable. Therefore the goal is to make an AI less predictable and more human-like.

\begin{figure}[b!]
  \centering
  \vspace{0pt}
  \includegraphics[width=0.75\linewidth]{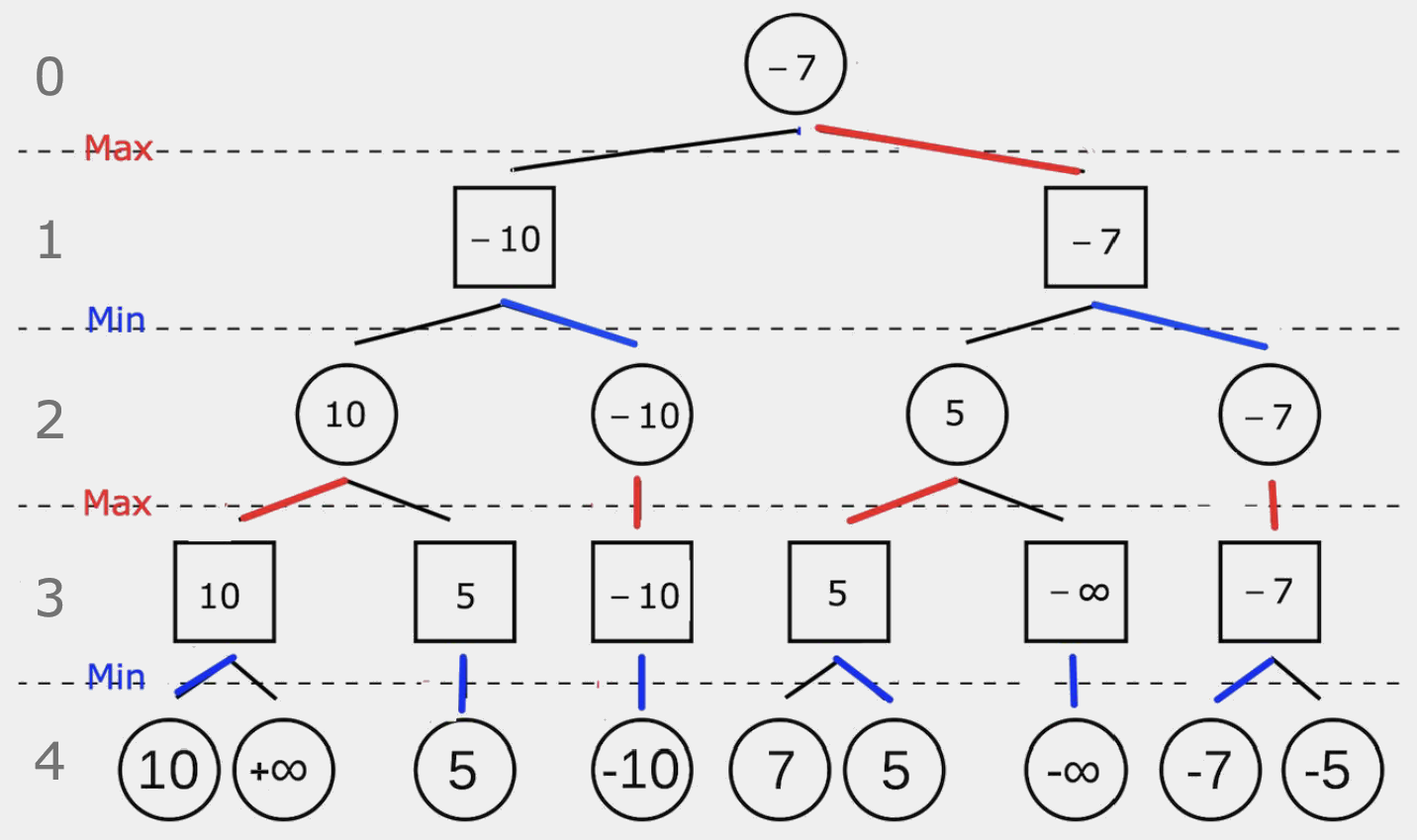}
  %\vspace{-20pt}
  \caption{Following the evaluation process in the Minimax algorithm, scores from various paths propagate upwards through the decision tree. The child node aligning with the score at the root node signifies the optimal path identified by the algorithm.}
  %\Description{}
  \label{fig:tree}
  \vspace{-15pt}
\end{figure}

\subsection{Implementation}
\subsubsection{Application}
Human decisions are often unpredictable, influenced by various factors such as personal emotions and environmental interruptions. This unpredictability poses a challenge for today's AI systems, which tend to perform poorly in such contexts. For instance, when faced with two identical situations, such as in a game, a conventional AI will consistently choose the same "best move". Our main objective is to enhance current AI algorithms to exhibit more unpredictability while still achieving specific goals. When modifying AI to incorporate randomness, it is essential to maintain the core objective: the AI must still aim to succeed, such as winning a game. Therefore, the fundamental algorithms should either remain unchanged or be only partially modified to incorporate random functions. In line with this principle, our work begins with traditional AI methods, focusing on the well-known Gobang game \cite{gobang}, where the goal is to achieve "five in a row." This game implemented in python will serve as a testbed for developing and evaluating our approach to making AI behavior more unpredictable while maintaining its effectiveness. We have chosen the application of a game environment, as it is a safe space, unlike an AI assistive self-driving car application.
%Particularly an emotional influence can drive decision-making in both beneficial and harmful ways [29]. This is apparent for the car-driving scenario. This also applies in the context of games of chance or gambling [30], here a rational investment should be lower than the expectation. However, people tend to spend money on activities like lotteries and insurance, which have negative expectations. Daniel Bernoulli pointed out that people's judgments when making decisions do not depend solely on the expectation itself [31]. Additionally, some researchers have proposed "rank and sign dependent theories," suggesting that the weight of a decision changes according to the probability and outcomes of the results [32].

\subsubsection{Minimax Algorithm}
The Gobang AI agent is fundamentally based on the Minimax and Alpha-Beta Pruning algorithms, commonly used in zero-sum games like chess. The Minimax Algorithm, particularly effective in zero-sum scenarios where one player's gain is another's loss, underpins the AI's strategy in Gobang. In such games, potential outcomes are represented in a decision tree, with each node assigned a score indicating the result's utility. For instance, in a simplified tree for Tic-Tac-Toe, the AI (MAX) aims to maximize its score (+1 for a win) while the opponent (MIN) seeks to minimize it (-1 for a loss). This structured approach ensures that the AI consistently strives for the best possible outcome.

The Gobang AI agent plays as MAX aiming to maximize benefits by exploring the entire decision tree depth-first and selecting paths with maximum scores. Conversely, the MIN agent minimizes benefits. Each node in the decision tree represents a game state, with scores propagated up from leaf nodes to determine optimal moves (see figure \ref{fig:tree}). In practice, the AI, represented as MAX in this scenario, evaluates potential moves several steps ahead to strategize its current best move based on game rules and board positions. 

\begin{figure}[h!]
  \centering
  \vspace{0pt}
  \includegraphics[width=\linewidth]{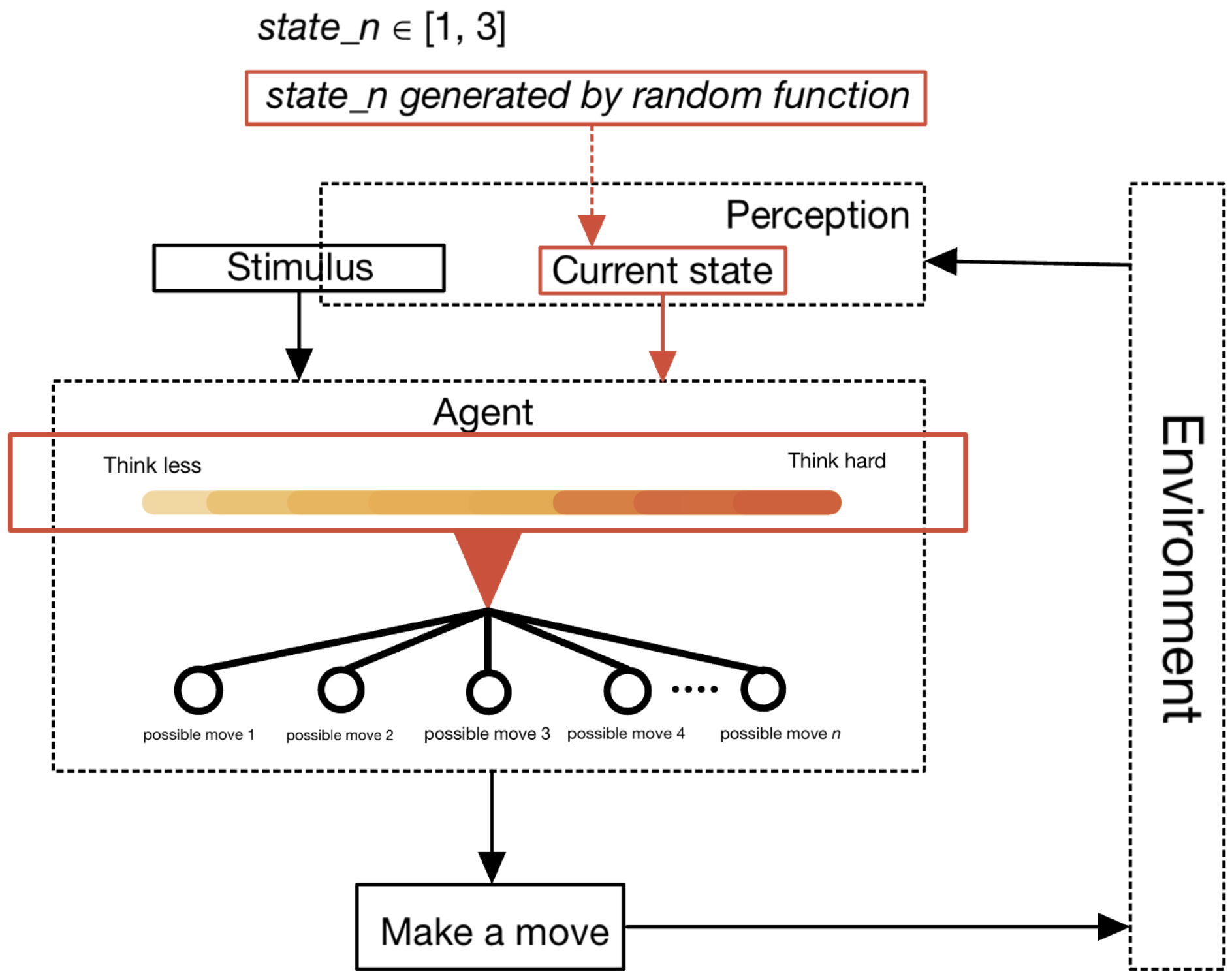}
  \vspace{-10pt}
  \caption{Injection Method A: The adjustment of thinking depth in the agent involves assigning an integer value from a random function to represent the agent's current mental state. A lower integer value, closer to 1, indicates a more carefree state, prompting the agent to perform a shallow tree search with a depth of 1 (Think Less). Conversely, higher integer values indicate a more cautious state, prompting a deeper tree search (Think More). The search yields a result based on this depth configuration.}
 % \Description{}
  \label{fig:meth1}
  \vspace{0pt}
\end{figure}

\subsubsection{Injection Method A}
The implementation of the first idea is straightforward. The simulated "emotional state" of the AI agent dictates the depth of its decision-making process, similar to a mapping relationship where each state corresponds to a specific search depth during iterations. 

\small
\begin{verbatim}
function minimax (node, depth, maximizingPlayer, state):
  if state == state n
     return state n mapped thinking depth n
  if depth n == 0 or node is a terminal node then
     return static evaluation of node
  if MaximizingPlayer
     maxEva = -infinity
     for each child of node do
        eva = minimax (child, depth n - 1, false, state n)
        maxEva = max (maxEva, eva) 
        return maxEva
  else
     minEva = +infinity
     for each child of node do
        eva = minimax (child, depth n - 1, true, state n)
        minEva = min (minEva, eva)
        return minEva
\end{verbatim}
\normalsize
    
In the pseudo code, the parameter "$state$" represents the simulated "emotional state" derived from our bionic random function, where "$state_n$" and "$depth_n$" form a pair in the state mapping table. Analogous to human behavior, a cautious or serious state prompts the agent to conduct deeper searches, whereas an imprudent state results in shallower considerations. The procedure outlined in the figure details this approach: the agent's current mental state, denoted by "$state_n$," translates into a numeric range [1 to 3] reflecting the extent of caution. Following the completion of the designated search depth, the agent selects the move yielding the highest score, thereby determining its next action. This method is also illustrated in figure \ref{fig:meth1}.

\begin{figure}[t!]
  \centering
  \vspace{0pt}
  \includegraphics[width=\linewidth]{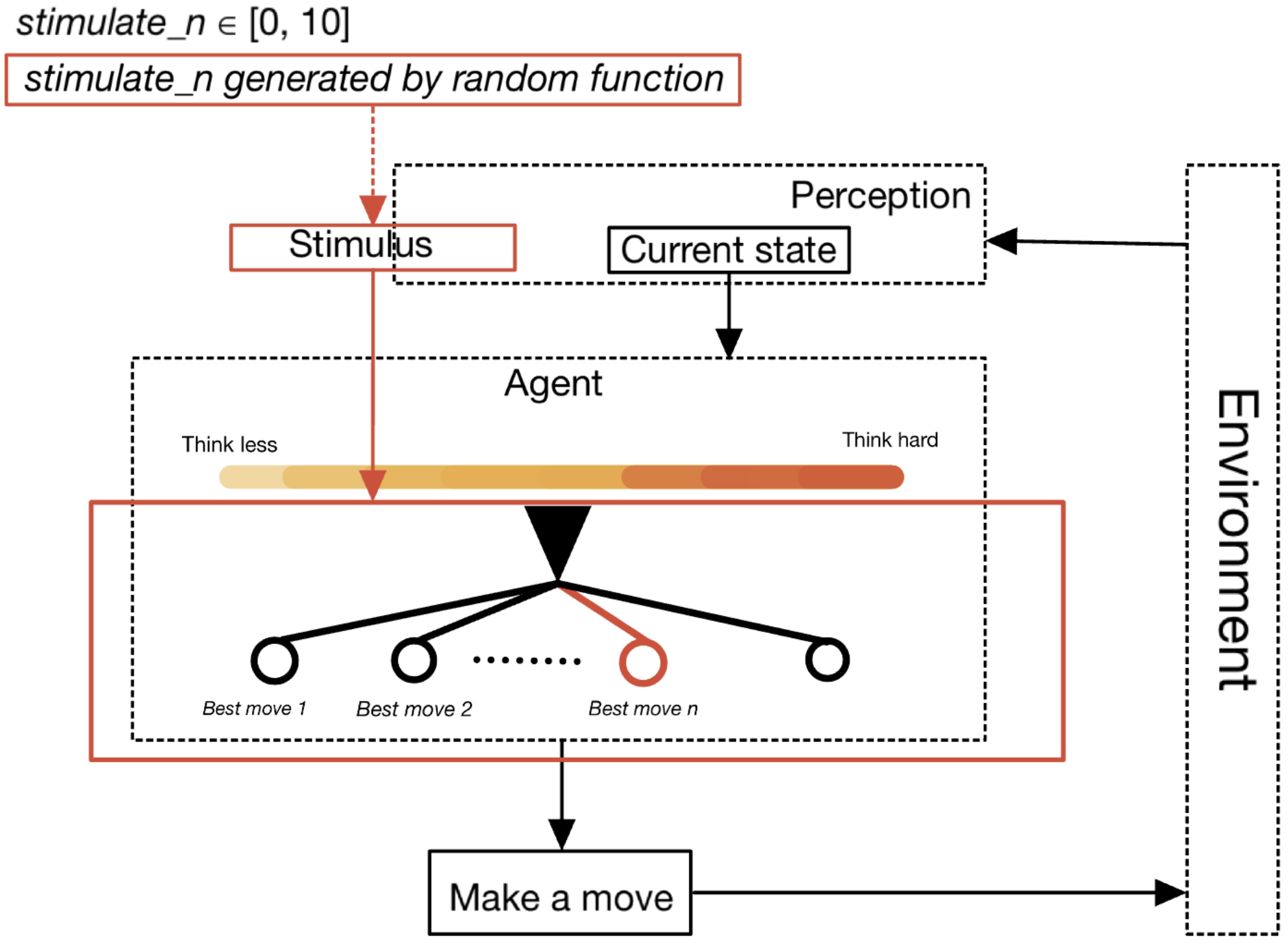}
  \vspace{-10pt}
  \caption{Injection Method B: The procedure involves providing stimulation to the agent after conducting the tree search. Once an integer representing the level of stimulation is inputted into the agent, it selects from the available moves. Higher levels of stimulation increase the likelihood that the agent will opt for less optimal moves—those with lower scores.}
  %\Description{}
  \label{fig:meth2}
  \vspace{0pt}
\end{figure}

\subsubsection{Injection Method B}
The implementation of the second idea involves a more intricate process. To enable our AI agent to select among various potential moves during its turn, we create an ordered list containing all possible moves. Subsequently, the agent's decision-making process is driven by an input stimulus that influences the selection from this list of moves. The pseudo-code for this approach is as follows:

\small
\begin{verbatim}
function minimax (node, depth, maximizingPlayer, stimulate):
  if stimulate == stimulate n
     return stimulate n mapped tendency to choose choice n
  new scorelist []
  if depth n as 0 or node is a terminal node then
     return static evaluation of node
  if MaximizingPlayer
     maxEva = -infinity
     for each child of node do
        eva = minimax (child, depth_n - 1, false, stimulate_n)
        if maxEva < eva then
           scorelist.add (eva)
           maxEva = eva
     scorelist.ordered (MaxToMin)
     return scorelist [choice n]
  else 
     minEva = +infinity
     for each child of node do
        eva = minimax (child, depth _n - 1, true, stimulate_n)
        if minEva > eva then
           scorelist.add (eva)
           minEva = eva
     scorelist.ordered (MinToMax)
     return scorelist[choice n1]
\end{verbatim}
\normalsize

Similar to the "$state$" parameter in the previous method, the "$stimulate$" input is also generated by our bionic random function. In this case, an integer ranging from 0 to 10 is chosen to represent varying degrees of stimulation. A higher stimulation value increases the likelihood that the AI may opt for a suboptimal move. This parameter may mimic an impulsive human decision-making—where curiosity or mood may influence risk-taking behavior. External interruptions during gameplay can also affect decision-making, similar to the influences represented by the random function in our agent's environment. Assuming the scores of each route equate to their winning probabilities, these are stored in "$scorelist$". For efficiency, not all routes need preservation; obviously disadvantageous ones can be discarded. Based on the specific "$stimulate$" value, denoted as "$stimulate_n$", our agent determines which routes to consider. In the subsequent figure \ref{fig:meth2}, the agent selects the "$best move n$" based on the "$stimulate_n$" input.

A common yet noteworthy scenario arises when two or more routes in the decision-making process yield identical or very similar scores. For instance, in the aforementioned figure \ref{fig:meth2}, "\emph{Best move 1}" and "\emph{Best move 2}" could correspond to different positions but share identical scores. In such instances, the choice between these routes has minimal impact on the current stage's likelihood of winning. To emulate human decision-making, our agent resorts to chance—randomly selecting one of these options to make its move.

\begin{figure*}[t]
\hspace{-83pt}
\begin{minipage}{7in}
 \vspace{-5pt}
  \centering
  \includegraphics[width=\linewidth]{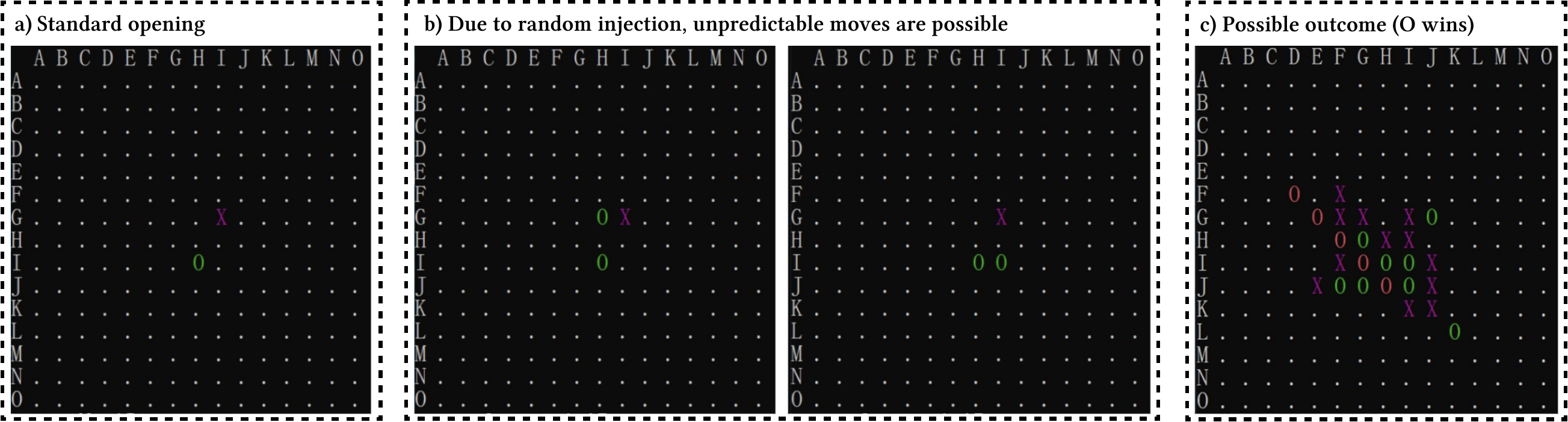}
  \vspace{-20pt}
  \caption{Pink X: original AI agent, Green O: our modified AI agents. a) Showing a standard opening. b) Our agent is able to continue with unpredictable moves due to the RNG inject. c) Five connecting chars in a row win the game -- our AI agent won.}
 % \Description{}
  \label{fig:gobang}
  \vspace{-10pt}
  \end{minipage}
\end{figure*}

\subsection{Evaluation}
It would be interesting to know whether the human would find playing against a bionic AI more unpredictable and difficult than playing against a simple rule-based AI agent, which is already provided by the Gobang Python implementation \cite{gobang} (see figure \ref{fig:gobang}). This rather subjective opinion may or may not arise in the player. The issue is that a non-experienced player is likely to lose against against a simple rule-based and bionic AI, as we quickly figured out in our pilot tests. We assume that only professional players may identify the difference of the varying game styles introduced by the different AIs in this particular application.

\subsubsection{Study Design}
Hence, we decided to utilize the provided rule-based AI as a benchmark, and testing several other modified copies against it. We calculated the win-rate based on 10 trails run in 50 blocks, resulting in 500 games played against every agent (2000 played games in total). To understand the strength of the modified agents, we additionally we asked 11 players to play against the standard AI 10 times, resulting in 110 played games.

\textbf{Agent A} is a copy of the original AI with the difference of using the Injection Method A, where we use the Python's random function (\textbf{pRNG}). \textbf{Agent B} relies on the Injection Method B, also using \textbf{pRNG}. \textbf{Agent C} combined both injection methods A+B using \textbf{pRNG}. \textbf{Agent Fly} combined both injection methods A+B using our bionic random dunction (\textbf{bRNG1}).

\subsubsection{Results}

\textbf{Agent C}, which exhibits random injections at several points results in a somewhat comparably low winning rate ($M$=0.34) to our amateur human players ($M$=0.39). Both are still able to win, however, inserting white noise at so many points into the \textbf{Agent C}, impacts the decision-making process negatively. \textbf{Agent A} ($M$=0.61) and \textbf{Agent B} ($M$=0.59) have just a single point of noise injection and surprisingly perform slightly better than the original AI. However, we also see a high standard deviation ($SD$=0.21) with a peak win-rate between 0.84 and 0.15. However, we can still see that inserting random noise is not a disadvantage and is somewhat confusing the original AI. Looking at the \textbf{Agent Fly}, we can see the highest win-rate ($M$=0.68) with a relatively low standard deviation ($SD$=0.12). Here, the bionic RNG inserted specific noise patterns that could be interpreted as a strategy from a human point of view. Apparently, the original AI was unable to counter this unpredictable behavior and therefore demonstrated the greatest loss against our bionic AI agent. 

\begin{table}[h!]
\centering
\vspace{-5pt}
  \caption{Average win-rates of different agents playing against the original AI. A high win-rate > 0.5 indicates the opponents, such as the Agent A, B... to have a greater chance to win against the original AI agent.}
  \vspace{-5pt}
  \label{tab:freq}
  \resizebox{0.685\textwidth}{!}{%
  \begin{tabular}{c|c|c|c|c|c}
    \toprule
  AI vs. & \textbf{Human} & \textbf{Agent A} & \textbf{Agent B} &\textbf{Agent C} & \textbf{Agent Fly}\\
    \midrule
    Win-Rate  & 0.39 & 0.61 & 0.59 & 0.34 & 0.68\\
  \bottomrule
\end{tabular}
}
\vspace{-10pt}
\end{table}

\section{Discussion}
%\subsection{Insights}

\emph{Bionic Random Function Generator (bRNG):}
%In general, the adjective "bionic" implies a combination of biological principles and electronic or mechanical technology. So, a "bionic random function", could theoretically be used to refer to a random function that is inspired by or mimics biological processes. While the term "bionic random function" isn't commonly used in scientific literature or technology, there is a concept that somewhat aligns with with it: a true random number generator (TRNG) \cite{wan2022flexible}. This concept is inspired by the uniqueness and randomness of biological architectures of a leaf, which is pointed through with a laser in order to achieve random number generation. While such device can be quite small, it is different to a living organism that has complex feelings and mood swings, which are somewhat reflected in our bRNG.
%
In essence, the term "bionic" typically denotes a fusion of biological principles with mechanical or electronic technology. Therefore, a "bionic random function", could theoretically be used to refer to a random function that is inspired by or mimics biological processes. While the term "bionic random function" isn't commonly used in scientific literature or technology, there is a concept that somewhat aligns with with it: a true random number generator (TRNG) \cite{wan2022flexible}. This concept draws inspiration from natural phenomena such as the unique and random architectures found in leaves, where a laser pointing through the tissue and an optical sensor is used to generate random numbers. In contrast, our bionic random number generator (bRNG) incorporates responses from a living organism, reflecting the complexity of emotions and mood variations.

\emph{Practical and Ethical Challenges:}
While the concept of leveraging a living organism raises intriguing ethical considerations, we chose an insect and designed an environment to mitigate many potential issues. Our aim was to create a habitat that is sufficiently spacious and provides adequate food to ensure the fly's well-being. However, our current setup has practical limitations: the device is bulky, and the process of triggering the fly's response via a fan activation may not be rapid enough for real-time applications. While effective for prototyping purposes, a miniaturized and portable device would be advantageous for practical implementation in various applications.

\emph{System's Robustness:}
Obtaining a bionic random value isn't always guaranteed due to the fly's occasional periods of rest when it might not respond to the fans. Moreover, the fly could sometimes hide in areas not fully captured by the camera, or go unnoticed if the YOLOv5s model, despite its high accuracy (91.4\%), missed identifying it.

\emph{A Quick and Dirty Approach:} 
We employed a powerful machine learning approach to accomplish a straightforward task: identifying the fly and tracking its flight vectors and coordinates for number generation. Alternatively, a simpler method could involve using a histogram of the camera image, which would enhance robustness and reduce computational costs.

\emph{Mission Accomplished:}
We confirmed that our bRNG, incorporating the fly's responses, significantly enhanced the AI agent, making it more unpredictable and successful. In comparison, both a human player and a conventional AI agent, even when enhanced with a white-noise RNG, exhibited lower performance levels.

\emph{Application in Game AI and Beyond:}
We initially tested our approach in the Gobang game, implementing our bionic RNG with the Minimax Algorithm to enhance decision-making unpredictability. This demonstrates significant potential, particularly also for other games such as chess. Adding unpredictability in various other AI applications, including LLMs, humanized robots, etc. could be a game changer particularly when it comes to making AI creative. Additionally, using a bRNG to generate cryptographic keys presents a promising avenue to bolster cybersecurity.

\emph{The Danger:}
This unpredictability poses risks, particularly in critical applications such as healthcare and finance, where reliability is essential. To balance these benefits and risks, unpredictability should be introduced in specific areas and in a controlled manner only. Although it is a necessary step in humanizing AI, it can be dangerous, as we might train AI to make mistakes, which might need to be covered up by an LLM creating lies.

\section{Conclusion}
%In conclusion, we believe that the integration of responses from a living fly into the decision-making process of AI represents a groundbreaking advancement in the rising field of bionic AI. 
%Through a comprehensive study comparing the performance of players against the bio-hybrid AI, standard AI, and random function, significant insights have been gained into the potential of this novel approach to enhance unpredictability and adaptability in AI systems. The methodology developed for creating bio-hybrid AI not only showcases its effectiveness in improving AI unpredictability but also opens up new avenues for exploring the intersection of biological and artificial systems. This research highlights the importance of incorporating diverse and unconventional sources of inspiration to push the boundaries of AI capabilities and foster innovation in decision-making processes. As we continue to bridge the gap between biological and artificial intelligence, the bio-hybrid AI approach offers a promising pathway towards developing more dynamic and nuanced AI systems that can better emulate the complexity and adaptability of human cognition. Overall, this study underscores the transformative potential of Bio-Hybrid AI in revolutionizing the future of artificial intelligence research and applications.

In this research, we presented a novel bionic AI system that leverages the natural, unpredictable responses of a living fly to enhance the decision-making unpredictability of an AI agent in the game of Gobang. Traditional AI systems, despite their reliability and precision, often fall short in replicating the nuanced and sometimes erratic decision-making processes characteristic of humans. By integrating a fly's diverse reactions into the AI's decision-making process, we introduced a level of unpredictability that mimics human-like spontaneity. Our comparative study showcased that the bionic AI agent not only outperformed human players but also surpassed conventional AI agents and those enhanced with white-noise randomization. This suggests that incorporating biologically inspired randomness can significantly improve AI performance in tasks that benefit from unpredictability.
 \bibliographystyle{splncs04}
 \bibliography{references.bib}

\begin{thebibliography}{10}
\providecommand{\url}[1]{\texttt{#1}}
\providecommand{\urlprefix}{URL }
\providecommand{\doi}[1]{https://doi.org/#1}

\bibitem{braiek2024machine}
Braiek, H.B., Khomh, F.: Machine learning robustness: A primer. arXiv preprint arXiv:2404.00897  (2024)

\bibitem{brown2020random}
Brown, J., Zhang, J.F., Zhou, B., Mehedi, M., Freitas, P., Marsland, J., Ji, Z.: Random-telegraph-noise-enabled true random number generator for hardware security. Scientific reports  \textbf{10}(1),  17210 (2020)

\bibitem{brundage2018malicious}
Brundage, M., Avin, S., Clark, J., Toner, H., Eckersley, P., Garfinkel, B., Dafoe, A., Scharre, P., Zeitzoff, T., Filar, B., et~al.: The malicious use of artificial intelligence: Forecasting, prevention, and mitigation. arXiv preprint arXiv:1802.07228  (2018)

\bibitem{chater2022paradox}
Chater, N., Zeitoun, H., Melkonyan, T.: The paradox of social interaction: Shared intentionality, we-reasoning, and virtual bargaining. Psychological Review  \textbf{129}(3), ~415 (2022)

\bibitem{fu2016alphago}
Fu, M.C.: Alphago and monte carlo tree search: the simulation optimization perspective. In: 2016 Winter Simulation Conference (WSC). pp. 659--670. IEEE (2016)

\bibitem{fuchs2022human}
Fuchs, T.: Human and artificial intelligence: A critical comparison. In: Intelligence-Theories and Applications, pp. 249--259. Springer (2022)

\bibitem{gaviria2017solution}
Gaviria~Rojas, W.A., McMorrow, J.J., Geier, M.L., Tang, Q., Kim, C.H., Marks, T.J., Hersam, M.C.: Solution-processed carbon nanotube true random number generator. Nano letters  \textbf{17}(8),  4976--4981 (2017)

\bibitem{greenblatt2023ai}
Greenblatt, R., Shlegeris, B., Sachan, K., Roger, F.: Ai control: Improving safety despite intentional subversion. arXiv preprint arXiv:2312.06942  (2023)

\bibitem{hinton2012improving}
Hinton, G.E., Srivastava, N., Krizhevsky, A., Sutskever, I., Salakhutdinov, R.R.: Improving neural networks by preventing co-adaptation of feature detectors. arXiv preprint arXiv:1207.0580  (2012)

\bibitem{jiang2017novel}
Jiang, H., Belkin, D., Savel’ev, S.E., Lin, S., Wang, Z., Li, Y., Joshi, S., Midya, R., Li, C., Rao, M., et~al.: A novel true random number generator based on a stochastic diffusive memristor. Nature communications  \textbf{8}(1), ~882 (2017)

\bibitem{johnson2020designing}
Johnson, J.: Designing with the mind in mind: simple guide to understanding user interface design guidelines. Morgan Kaufmann (2020)

\bibitem{jones2024ai}
Jones, N.: Ai now beats humans at basic tasks—new benchmarks are needed, says major report. Nature  \textbf{628}(8009),  700--701 (2024)

\bibitem{kasper2021detecting}
Kasper-Eulaers, M., Hahn, N., Berger, S., Sebulonsen, T., Myrland, {\O}., Kummervold, P.E.: Detecting heavy goods vehicles in rest areas in winter conditions using yolov5. Algorithms  \textbf{14}(4), ~114 (2021)

\bibitem{katanforoosh2019initializing}
Katanforoosh, K., Kunin, D.: Initializing neural networks. DeepLearning. ai  (2019)

\bibitem{korteling2021human}
Korteling, J.H., van~de Boer-Visschedijk, G.C., Blankendaal, R.A., Boonekamp, R.C., Eikelboom, A.R.: Human-versus artificial intelligence. Frontiers in artificial intelligence  \textbf{4},  622364 (2021)

\bibitem{kukreja2016introduction}
Kukreja, H., Bharath, N., Siddesh, C., Kuldeep, S.: An introduction to artificial neural network. Int J Adv Res Innov Ideas Educ  \textbf{1}(5),  27--30 (2016)

\bibitem{lebovitz2021ai}
Lebovitz, S., Levina, N., Lifshitz-Assaf, H.: Is ai ground truth really true? the dangers of training and evaluating ai tools based on experts'know-what. MIS quarterly  \textbf{45}(3) (2021)

\bibitem{li2021random}
Li, X., Zanotti, T., Wang, T., Zhu, K., Puglisi, F.M., Lanza, M.: Random telegraph noise in metal-oxide memristors for true random number generators: A materials study. Advanced Functional Materials  \textbf{31}(27),  2102172 (2021)

\bibitem{manoonpong2021insect}
Manoonpong, P., Patan{\`e}, L., Xiong, X., Brodoline, I., Dupeyroux, J., Viollet, S., Arena, P., Serres, J.R.: Insect-inspired robots: bridging biological and artificial systems. Sensors  \textbf{21}(22), ~7609 (2021)

\bibitem{mcculloch1943logical}
McCulloch, W.S., Pitts, W.: A logical calculus of the ideas immanent in nervous activity. The bulletin of mathematical biophysics  \textbf{5},  115--133 (1943)

\bibitem{mishra2014view}
Mishra, M., Srivastava, M.: A view of artificial neural network. In: 2014 international conference on advances in engineering \& technology research (ICAETR-2014). pp.~1--3. IEEE (2014)

\bibitem{lotto2015}
Nestel, M.L.: Inside the biggest lottery scam ever. The Daily Beast  (2015), \url{https://www.thedailybeast.com/articles/2015/07/07/inside-the-biggest-lottery-scam-ever}

\bibitem{noll1998method}
Noll, L.C., Mende, R.G., Sisodiya, S.: Method for seeding a pseudo-random number generator with a cryptographic hash of a digitization of a chaotic system (Mar~24 1998), uS Patent 5,732,138

\bibitem{oritsegbemi2023human}
Oritsegbemi, O.: Human intelligence versus ai: Implications for emotional aspects of human communication. Journal of Advanced Research in Social Sciences  \textbf{6}(2),  76--85 (2023)

\bibitem{ouyang2019vehicle}
Ouyang, L., Wang, H.: Vehicle target detection in complex scenes based on yolov3 algorithm. In: IOP Conference Series: Materials Science and Engineering. vol.~569, p. 052018. IOP Publishing (2019)

\bibitem{pan2017virtual}
Pan, X., You, Y., Wang, Z., Lu, C.: Virtual to real reinforcement learning for autonomous driving. arXiv preprint arXiv:1704.03952  (2017)

\bibitem{phillips2008intelligent}
Phillips-Wren, G., Ichalkaranje, N.: Intelligent decision making: An AI-based approach, vol.~97. Springer Science \& Business Media (2008)

\bibitem{quinlan1996learning}
Quinlan, J.R.: Learning decision tree classifiers. ACM Computing Surveys (CSUR)  \textbf{28}(1),  71--72 (1996)

\bibitem{redmon2016you}
Redmon, J., Divvala, S., Girshick, R., Farhadi, A.: You only look once: Unified, real-time object detection. In: Proceedings of the IEEE conference on computer vision and pattern recognition. pp. 779--788 (2016)

\bibitem{romano2019review}
Romano, D., Donati, E., Benelli, G., Stefanini, C.: A review on animal--robot interaction: from bio-hybrid organisms to mixed societies. Biological cybernetics  \textbf{113},  201--225 (2019)

\bibitem{segovia2022revisiting}
Segovia-Cu{\'e}llar, A.: Revisiting the social origins of human morality: A constructivist perspective on the nature of moral sense-making. Topoi  \textbf{41}(2),  313--325 (2022)

\bibitem{song2015decision}
Song, Y.Y., Ying, L.: Decision tree methods: applications for classification and prediction. Shanghai archives of psychiatry  \textbf{27}(2), ~130 (2015)

\bibitem{sutton2018reinforcement}
Sutton, R.S., Barto, A.G.: Reinforcement learning: An introduction. MIT press (2018)

\bibitem{swaroop2024accuracy}
Swaroop, S., Bu{\c{c}}inca, Z., Gajos, K.Z., Doshi-Velez, F.: Accuracy-time tradeoffs in ai-assisted decision making under time pressure. In: Proceedings of the 29th International Conference on Intelligent User Interfaces. pp. 138--154 (2024)

\bibitem{talati2024ai}
Talati, D.: Ai (artificial intelligence) in daily life. Authorea Preprints  (2024)

\bibitem{thuan2021evolution}
Thuan, D.: Evolution of yolo algorithm and yolov5: The state-of-the-art object detention algorithm  (2021)

\bibitem{urmson2008autonomous}
Urmson, C., Anhalt, J., Bagnell, D., Baker, C., Bittner, R., Clark, M., Dolan, J., Duggins, D., Galatali, T., Geyer, C., et~al.: Autonomous driving in urban environments: Boss and the urban challenge. Journal of field Robotics  \textbf{25}(8),  425--466 (2008)

\bibitem{wan2022flexible}
Wan, Y., Chen, K., Huang, F., Wang, P., Leng, X., Li, D., Kang, J., Qiu, Z., Yao, Y.: A flexible and stretchable bionic true random number generator. Nano Research  \textbf{15}(5),  4448--4456 (2022)

\bibitem{WANG20104052}
Wang, X., Zhao, J.: An improved key agreement protocol based on chaos. Communications in Nonlinear Science and Numerical Simulation  \textbf{15}(12),  4052--4057 (2010). \doi{https://doi.org/10.1016/j.cnsns.2010.02.014}, \url{https://www.sciencedirect.com/science/article/pii/S1007570410001103}

\bibitem{wen2021advanced}
Wen, C., Li, X., Zanotti, T., Puglisi, F.M., Shi, Y., Saiz, F., Antidormi, A., Roche, S., Zheng, W., Liang, X., et~al.: Advanced data encryption using 2d materials. Advanced Materials  \textbf{33}(27),  2100185 (2021)

\bibitem{yampolskiy2019monitorability}
Yampolskiy, R.V.: Unpredictability of ai. AI and Ethics pp. 1--19 (2019), \url{https://arxiv.org/pdf/1905.13053}

\bibitem{gobang}
Zhou, R.: Gobang game with artificial intelligence in 900 lines !! (2015), \url{https://github.com/142857why/gobang-python}

\end{thebibliography}

\end{document}